\documentclass[11pt,a4paper]{article}
\usepackage{graphicx}
\usepackage{hyperref}
\usepackage{subcaption}
\usepackage{mathtools}

\newcommand*{\tp}{\mathrm{TP}}
\newcommand*{\fp}{\mathrm{FP}}
\newcommand*{\tn}{\mathrm{TN}}
\newcommand*{\fn}{\mathrm{FN}}

\renewcommand{\alpha}{\theta}
\newcommand*{\mc}[1]{\mathcal{#1}}
\newcommand*{\al}[1]{{\alpha_{#1}}}

\newcommand*{\R}{\mathds{R}}
\usepackage{dsfont}
\usepackage{amsmath,amsthm,amssymb}
\newcommand\newsubcap[1]{\phantomcaption%
       \caption*{\figurename~\thefigure\thesubfigure: #1}}

\begin{document}
\title{Random 2.5D U-net for Fully 3D Segmentation}

\author{Christoph Angermann and  Markus Haltmeier\\ \\ 
	Department of Mathematics, University of Innsbruck\\ Technikerstra{\ss}e 13, A-6020 Innsbruck\\
\tt\{christoph.angermann,markus.haltmeier\}@uibk.ac.at\\
\url{http://applied-math.uibk.ac.at}}

\maketitle   
    
\begin{abstract}
Convolutional neural  networks are state-of-the-art  
for various segmentation tasks.  
While for 2D images these networks are also computationally  efficient, 
3D  convolutions  have  huge storage requirements and therefore, end-to-end training is limited by GPU memory and data size.
To overcome this issue,   we introduce a network structure for volumetric data 
 without 3D  convolution layers. 
 The main idea is to include projections from different  directions  to transform the volumetric 
  data to a sequence of   images, where each image contains 
 information of the full data.
 We then apply  2D convolutions  to these projection images 
 and lift them again to volumetric data using a trainable reconstruction algorithm. 
 The proposed   architecture can be applied end-to-end to very large data volumes without cropping or sliding-window techniques. For a tested sparse binary segmentation task, it outperforms already known standard approaches and is  more resistant to generation of artefacts.
  
\end{abstract}

\section{Introduction}


Deep convolution neural networks have become a powerful method for image recognition \cite{KH16,vgg}. In the last few years they also exceeded the state-of-the-art in providing segmentation masks for 2D images. Long et al. \cite{JL15} proposed the idea to transform VGG-nets \cite{vgg} to deep convolution filters for obtaining semantic segmentations of 2D data. Based on these deep convolution filters, Ronneberger et al. \cite{OR15} introduced a novel network architecture, the so-called U-net.
With this architecture they redefined the state-of-the-art in image annotation till today. The U-net provides a powerful 2D segmentation tool for biomedical applications, since it has been demonstrated to learn highly accurate segmentation masks from only very few training samples.

Among others,  the  fully automated generation of volumetric segmentation masks is becoming more and more important for  biomedical applications.
This task  is still challenging. 
One idea is to extend the U-net structure 
to volumetric data by using 3D convolutions, as has been proposed in \cite{OC16,EB17,vnet}. Some special applications require end-to-end segmentation, where it is disadvantageous to use sliding-window approaches or to work with smaller patches. For such end-to-end tasks, significant drawbacks  of  3D convolution models  are the huge memory requirements and the resulting restrictions to the model's complexity. Deep learning segmentation  
methods  are therefore often applied to 2D slice images \cite{OC16}. However, 
these slice images do not contain  information of the full 3D data, which makes the segmentation task much more challenging.

To address the  drawbacks  of existing  approaches, 
we  introduce  a  network structure which is able to 
generate accurate  3D segmentation masks of very large volumes.  The main idea is  to include projection layers from different directions which transform the data to 2D images containing
information of the full 3D data. The projection method mainly depends on the application. Common used methods are the {maximum intensity projection} or the {Radon-transform} \cite{radon}. We then apply the 2D U-net to these projection images and propose a learnable reconstruction algorithm to lift them again to volumetric data.

\begin{figure}[htb!]
	\centering
	\begin{subfigure}{0.24\textwidth}
		
		\includegraphics[width=0.9\textwidth]{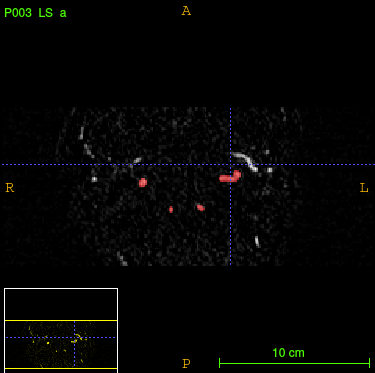}
		\subcaption{ Transversal.}
	\end{subfigure}
	\begin{subfigure}{0.24\textwidth}
		\includegraphics[width=0.9\textwidth]{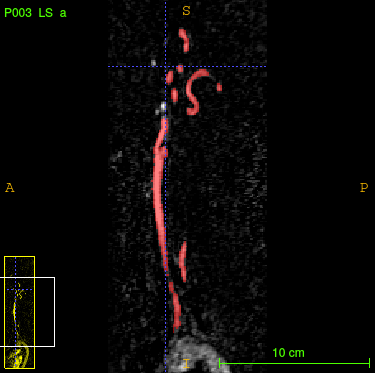}
		\subcaption{ Sagittal.}
	\end{subfigure}
	\begin{subfigure}{0.24\textwidth}
		\includegraphics[width=0.9\textwidth]{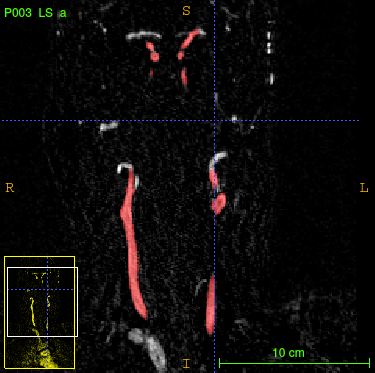}
		\subcaption{ Coronal.}
	\end{subfigure}
	\begin{subfigure}{0.24\textwidth}
		\includegraphics[width=0.89\textwidth]{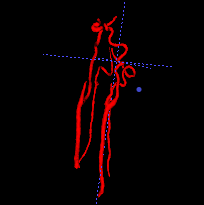}
		\subcaption{ 3D annotation.}
	\end{subfigure}
	\caption{ In every plane (a)-(c) the blood vessels of interest are marked in red. In (d) we see the corresponding 3D segmentation mask. The segmentation was conducted with the freeware ITK-SNAP \cite{ITK}.}
	\label{fig:fig3}
\end{figure}

As one targeted application, we test the proposed model for segmenting 
blood vessels   in  3D magnetic resonance 
angiography (MRA) scans (Fig. \ref{fig:fig3}).
We are given volumetric MRA scans of 119  patients provided by the Department of Neuroradiology Innsbruck \footnote{\url{https://www.i-med.ac.at/neuroradiologie}}, which face the arteries and veins between the brain and the chest. Also the 3D ground truths of these 119 patients have been provided.  These segmentation masks  have been generated by hand which takes hours for each patient.
Our goal  is the fully automated generation of  sparse 3D
segmentation masks of those vessels.   For that 
purpose we use deep learning and neural networks.
At the first glance, this problem may seem to be quite easy 
because we only have 
two labels (background and blood vessel), where  thresholding could be applied. However, we are not interested into segmentation of those vessels with highest intensity, but in those which  assist the doctor to detect dangerous to health abnormalities. This is the reason why we can not use  sliding-window techniques, since the model is not able to determine out of  a small patch if the seen vessel is  ``important'' for the segmentation. Other challenges  are caused by the big size of the 
volumes ($96 \times 288 \times 224$ voxels) and by the very unbalanced distribution (in average, 99.76 \% of all voxels indicate background).

In the following Section, we give a outline of some already existing methodologies and then propose the projection-based 2.5D U-net architecture in Section \ref{sec:2.5d}. In Section \ref{results}, we discuss some numerical results on our targeted application.
This work is mainly based on  research results shown in \cite{CA19}. Therefore, some parts are quite similar.

\section{Background}

\subsection{3D U-net}
\label{sec:u-net}

The U-net used for  binary 3D segmentation  is a mapping
$\mc{U} \colon \R^{a \times b \times c } \to [0,1]^{a \times b \times c } 
$  which takes 3D data scans  as input and outputs 
for each voxel the probability of being a foreground voxel (e.g. denoting a blood vessel). The 3D U-net used in this paper has exactly the same structure as in \cite{OR15}    with the use of 3D convolution and 3D pooling layers.  Due to the huge size of the training samples ($96\times288\times224$ 
voxels), we have to take special care of memory space if we want to conduct  end-to-end segmentation. This causes restrictions to the amount of downsampling steps and
channel sizes in the model.
Therefore, we are  able to conduct only  4 downsampling steps  with channel size 128 in the deepest convolution block to ensure, end-to-end training is still manageable by our \textit{NVIDIA GeForce RTX 2080} GPU.
 This results into
a model consisting of $1.6 \times 10^6$ trainable parameters, which are updated for every training sample (minibatch size 1). 
The model is trained with  the 
Dice-loss function \cite{vnet}  
\begin{equation}
\label{diceloss}
\ell(y,\hat y) = 1 - \frac{ 2 \sum_k (y  \odot \hat y)_k }{\sum_k \hat{y}_k+\sum_k y_k},
\end{equation}
where $\odot$ denotes  pointwise multiplication,  
the  sums are taken over all voxel locations, 
$\hat y = \hat{f}(x) $ are the probabilities predicted 
by the U-net,  and $y$ is the corresponding ground truth.

\subsection{Slice-by-Slice 2D U-Net}
The naive approach for  reducing memory requirements of 3D segmentation is to process each of the slice images independently through a 2D segmentation network (compare \cite{OC16}). 

For application, we take our MRA scans and extract their slices to a dataset, which now consists of over 10000 slice-images of size $288\times 224$. The next step is to train a 2D segmentation model like the 2D U-net  \cite[Fig. 1]{OR15} with approximately $8.6\times 10^6$ parameters. Here we also make  use of the Dice-loss function in Equation (\ref{diceloss}). Stacking these segmented slices to 3D volumes again after propagating them through the 2D network yields the 3D segmentation masks.

\section{Projection-Based 2.5D U-net}
\label{sec:2.5d}

\subsection{Proposed 2.5D U-net architecture}
 \label{sec:determ}
\begin{figure}
	\centering
	\includegraphics[width=0.75\textwidth]{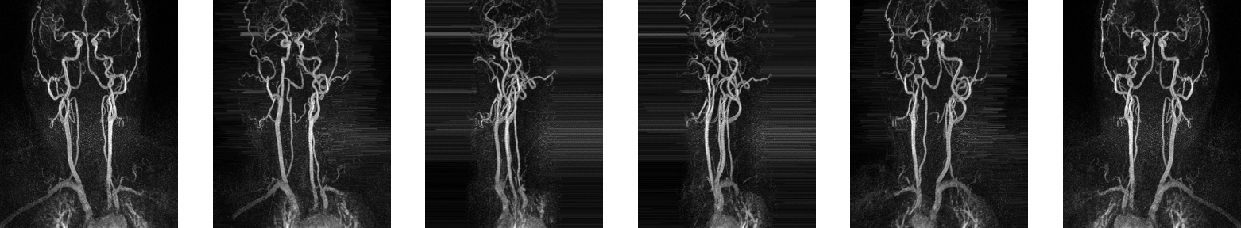}\\
	\includegraphics[width=0.75\textwidth]{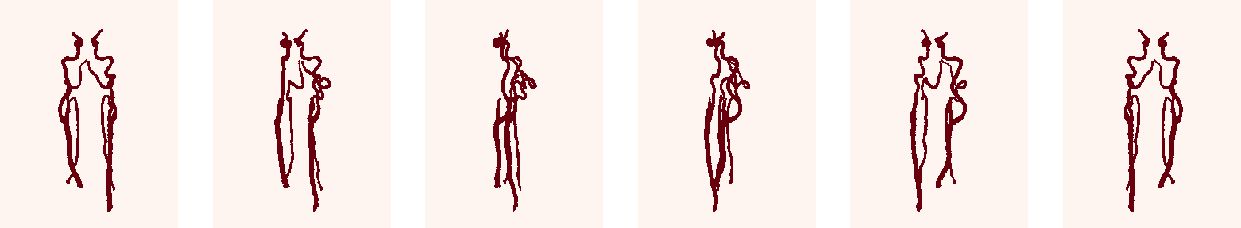}
	\caption{MIP images of a 3D MRA scan with directions  $\alpha\in\{k\times 36^\circ\mid k=0,\ldots,5\}$.}
	\label{fig:fig11}
\end{figure}

As mentioned in the introduction, the main idea is to include projection layers from different directions. Due to the high sparsity and the orientation of the vessels in our application, maximum intensity projection (MIP) as projection technique seems to be an appropriate choice (Fig. \ref{fig:fig11}).

\begin{figure}[h!!]
	\begin{subfigure}{0.5\textwidth}
		\centering	
		\includegraphics[width=0.845\textwidth]{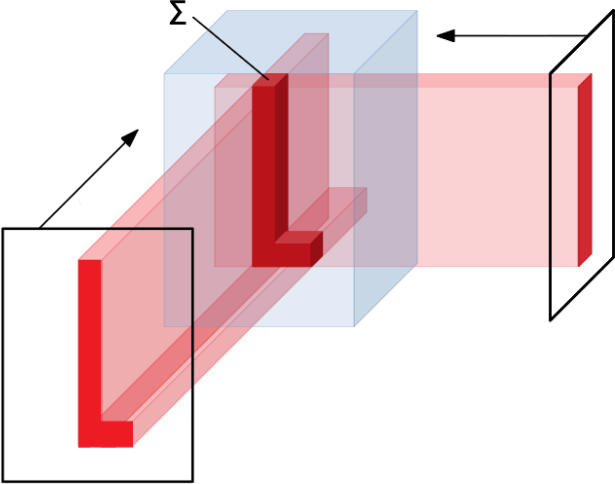}
		\newsubcap{ \textbf{Reconstruction operator} $\mc{R}_{2,\{0,90\}}$: Voxel value is defined as the sum over the corresponding 2D values.}
		\label{fig:fig18}
	\end{subfigure}\hspace{0.03\textwidth}
	\begin{subfigure}{0.48\textwidth}
		\centering
		\includegraphics[width=0.485\textwidth]{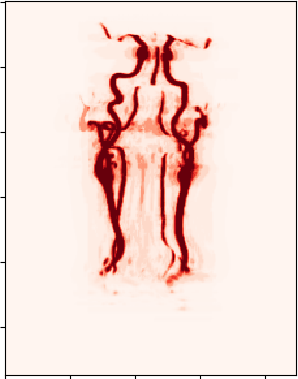}\hfil			\includegraphics[width=0.485\textwidth]{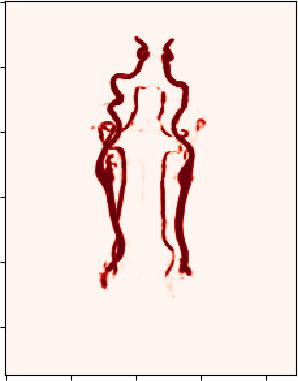}
		\newsubcap{Network's output  without filtration (left) and with filtration (right).}
		\label{fig:fig19}
	\end{subfigure}
\end{figure}

The proposed 2.5D U-net takes the form
\begin{align}\label{eq1}
\mc{N}({x})=\mc{T}\circ\mc{R}_{p,\Theta}\circ\mc{F}_p\circ
\begin{bmatrix}
\mc{U} \circ \mc{M}_{\alpha_1} ({x})
\\ \vdots \\ 
\mc{U} \circ \mc{M}_{\alpha_p} ({x})
\end{bmatrix},
\end{align}
where 
\begin{itemize}
	\item $\mc{M}_{\alpha_i}: \mathbb{R}^{a\times b\times c}\to \mathbb{R}^{b\times c}$ are MIP images for different directions $\al{1},\ldots,\al{p}$,
	\item $\mc{U}:\mathbb{R}^{b\times c}\to [0,1]^{b\times c}$ is exactly the same 2D U-net structure as in  \cite[Fig. 1]{OR15}, 
	\item $\mc{F}_p:\big([0,1]^{b\times c}\big)^p\to \big(\mathbb{R}^{b\times c}\big)^p$ is a learnable filtration,
	\item $\mc{R}_{p,\Theta}:\big(\mathbb{R}^{b\times c}\big)^p \to \mathbb{R}^{a\times b\times c}$ is a reconstruction operator using $p$ \textit{linear backprojections} for directions $\alpha_1,\ldots,\alpha_p\in\Theta$ as illustrated in Fig. \ref{fig:fig18},
	\item $\mc{T}:\mathbb{R}^{a\times b\times c} \to [0,1]^{a\times b\times c}$ is a fine-tuning operator (average pooling followed by a learnable  normalization  followed by the sigmoid activation).
\end{itemize}
The backprojection operator $\mc{R}_{p,\Theta}$ causes a  shadow (Fig. \ref{fig:fig19}, left), which suggests the usage of  filtrated backprojection. Therefore, we apply a convolution layer $\mc{F}_p$ before linear backprojection. Using  $1\times 2$ filters, which get adapted during training for each projection direction  $\alpha \in \Theta$ individually, solves the problem (Fig. \ref{fig:fig19}, right).
For the fine-tuning operator $\mc{T}$ we use average pooling with pool-size (2, 2, 2). This is followed by a learnable normalization, that additionally shifts the pooled data by an  adjusted parameter since the decision boundaries have been changed by the operator $\mc{R}_{p,\Theta}$. 

\subsection{Random 2.5D U-net}

\begin{figure}[h!]
	\centering
	\includegraphics[width=0.8\textwidth]{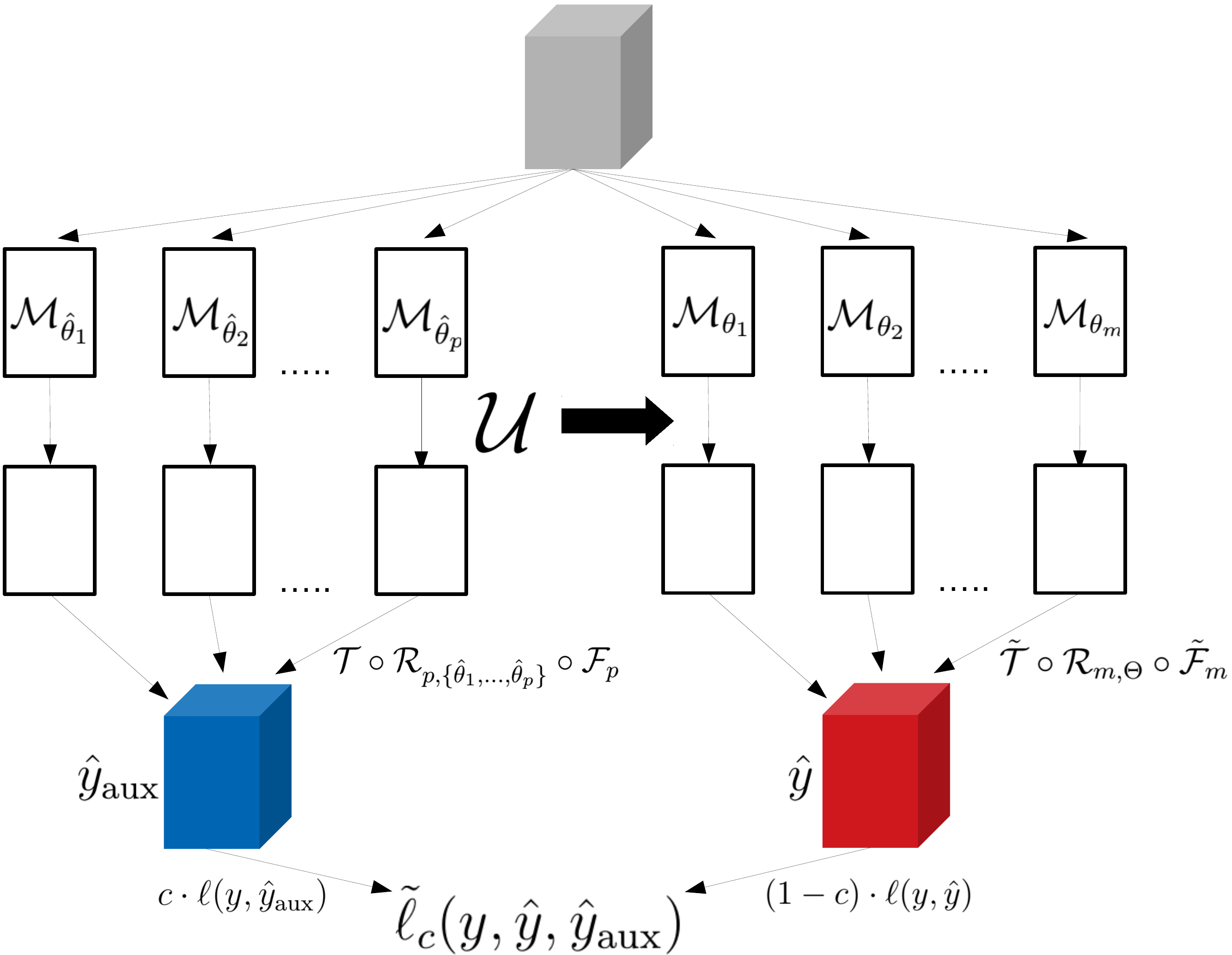}
	\caption{Architecture of the random 2.5D U-net.}
	
	\label{fig:fig34}
\end{figure}

We set ${\Theta} = \{k\times \frac{180}{m}\mid k=0,\ldots,m-1\}$ and construct MIP images for all $\alpha \in \Theta$, where $m$ denotes the number of projections we want to use for the volumetric reconstruction.
Again considering end-to-end training, it is not possible to update the  model  $\mc{N}$ using all $|\Theta|=m$ projection images if $m$ is large due to memory issues. Therefore, we propose a random training strategy with two paths:
\begin{enumerate}
	\item \textbf{Path 1:} 
	We consider for $k=0,\ldots,p-1,\ p\ll m,$ the set $\Theta_k \triangleq \{k,k+1,\ldots,k+\frac{180}{p}-1\}$
	and choose for each epoch one angle $\hat{\alpha}_k \in \Theta_k$ uniformly at random. For that $p$ projection directions, we generate the corresponding MIP images out of the 3D input $x$ and process them through the network $\mc N$, that has the same structure as in Equation (\ref{eq1}). The first output $\hat{y}_\text{aux}$ is then the learned reconstruction of the $p$ randomly chosen MIP segmentations. Note, that  the 2D convolution part $\mc U$ in Equation (\ref{eq1}) is also trained with these random projections. 
	
	\item \textbf{Path 2:} 
	Here we generate for all $m$ projection directions in $\Theta$ the corresponding MIP images out of the same 3D input $x$. We take the 2D convolution network $\mc U$ out of the first part and apply it to the $m$ projection images. Note, that  $\mc U$ is only needed for application and we do not update parameters of $\mc U$ in this path (Fig. \ref{fig:fig34}). Given the $m$ predictions $\mc U\big(M_{\theta_1}(x)\big),\ldots,	\mc U\big(M_{\theta_m}(x)\big)$, we are able to train a filtration layer $\tilde{\mc{{F}}}_m$ and a new fine-tuning operator $\tilde{\mc T}$. These are the only layers which get updated in this path. The final output of the second path is then computed by
	$
	\hat{y}=\tilde{\mc T}\circ \mc{R}_{m,\Theta} \circ \tilde{\mc{{F}}}_m \circ \big[\mc U\big(M_{\theta_1}(x)\big),\ldots,	\mc U\big(M_{\theta_m}(x)\big) \big].
	$
	Note that this output contains information of all $m$ projection directions.
	The final segmentation is generated by applying a threshold of 0.5 to $\hat y$.
\end{enumerate}

In conclusion, although  we only use $p$ random directions for adjustment of the 2D convolution part in $\mc{N}$ (first path), we are able with the help of $\mc U$ to construct simultaneously volumetric segmentation masks using available information from all $m$ projection directions $\alpha \in \Theta$ (second path) with $m\gg p$.
Experiments show, that $p=12$ in path 1 and $m=60$ in path 2 are  appropriate choices for our task, since for higher values only computational costs increase but not the  accuracy.

We make use of the following joint Dice-loss function:
\begin{align*}
\tilde{\ell}_c(y,\hat{y},\hat{y}_\text{aux})=c\cdot {\ell}(y,\hat{y}_\text{aux})+
(1-c)\cdot {\ell}(y,\hat{y}),
\end{align*}
where $y$ is the ground truth related to input tensor $x$, $\hat{y}_\text{aux}$ is the model's first output (path 1), $\hat{y}$ is the model's second output (path 2) and therefore the final segmentation prediction, and $\ell$ is the Dice-loss function in Equation (\ref{diceloss}).  We choose the regularization
$
c(n_\text{epoch}+1)=c(n_\text{epoch})\cdot 0.99^{n_\text{epoch}}
$ for balance parameter $c$,
where $c$ gets initialized with $c(1)=0.99$. 

\section{Numerical Results}
\label{results}

We  evaluate the following  metrics  during training:
\begin{itemize}
	\item \textit{mean accuracy} \cite{JL15}:\hspace{2.52cm}
	$
	\textbf{MA}\triangleq 
	\frac{1}{2} \left(\frac{\tn}{\tn+\fp}+\frac{\tp}{\tp+\fn}\right)
	$
	
	\item \textit{mean intersection over union} \cite{JL15}:\hspace{0.1cm}
	$
	\textbf{IU}\triangleq 
	\frac{1}{2} \left(\frac{\tn}{\tn+\fp+\fn}+\frac{\tp}{\tp+\fn+\fp}\right)$
	
	\item \textit{Dice-coefficient} \cite{EB17,vnet}:\hspace{2.05cm}
	$
	\textbf{DC} \triangleq 
	\frac{2\cdot\tp}{2\cdot\tp+\fn+\fp}$
\end{itemize}
For training,  Adam-optimizer \cite{adam} is used with learning rate 0.001 in combination with learning-rate-scheduling, i.e. if validation loss does not decrease within 3 epochs the learning rate gets halved. If the network shows no improvement for 5 epochs, the training process gets stopped (early stopping) and the weights of the best epoch in terms of validation loss get restored. For every model we conduct 7 training-runs using cross-validation:

\begin{itemize}
	\item \textbf{3D U-net:} Dice-loss of $\textbf{0.195}\pm\textbf{ 0.088}$, MA of $91.1\%\pm 2.82\%$,  IU of $86.3\%\pm 3.57\%$ and DC of $\textbf{82\%}\pm\textbf{ 6.07\%}$.	Average total training time  is  31.8 minutes on \textit{NVIDIA GeForce RTX 2080} GPU and  a memory space of 6.15 gigabyte is allocated during training. Application time to a test sample  is 235 ms.
	\item \textbf{slice-by-slice 2D U-net:} Dice-loss of $\textbf{0.154}\pm \textbf{0.095}$, MA of $91.8\%\pm 6.77\%$,  IU of $87.1\%\pm 5.86\%$ and DC of $\textbf{84.6\%}\pm \textbf{9.56\%}$. Total training time  is  16.8 minutes on GPU and  a memory space of 1.83 gigabyte is allocated. Application  time is 473 ms.
	\item \textbf{random 2.5D U-net:} Dice-loss of $\textbf{0.148}\pm \textbf{0.019}$, MA of $92.7\%\pm 1.21\%$,  IU of $87.6\%\pm 1.34\%$ and DC of $\textbf{85.6\%}\pm \textbf{1.91\%}$. Total training time  is  58.7 minutes on GPU and  a memory space of 6.69 gigabyte is allocated. Application time is 349 ms.
\end{itemize}
\begin{figure}[h!]
	\begin{subfigure}{0.47\textwidth}
		\centering
		\includegraphics[width=0.96\textwidth]{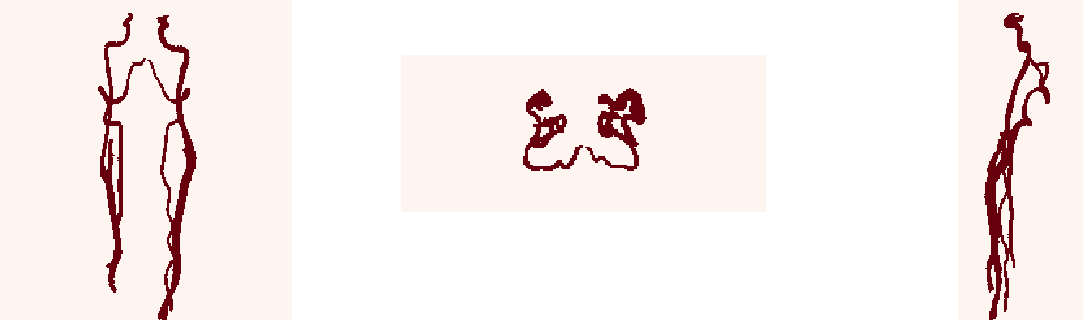}\\
		\includegraphics[width=0.96\textwidth]{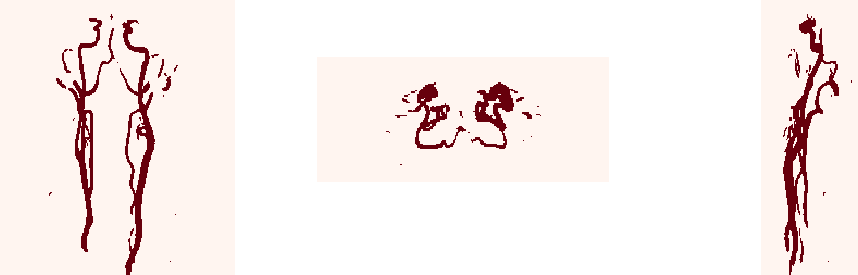}\\
		\includegraphics[width=0.96\textwidth]{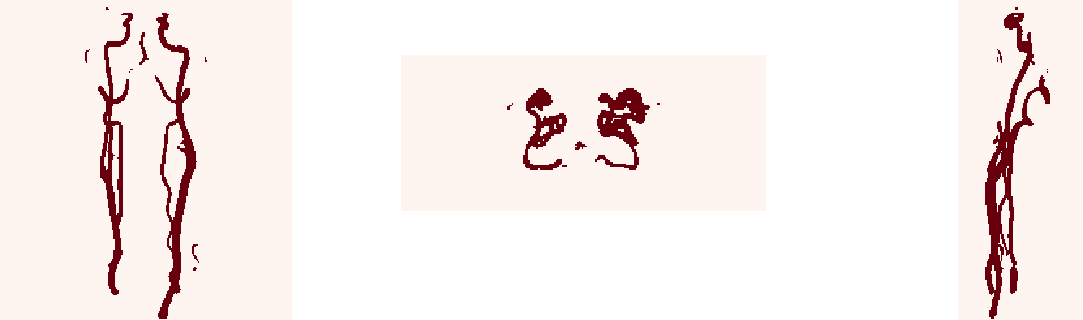}\\
		\includegraphics[width=0.96\textwidth]{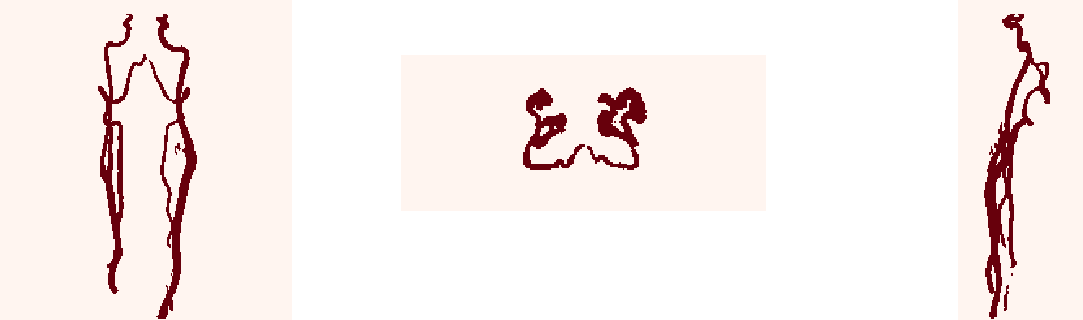}
		\newsubcap{Ground truth (first), segmentation generated by the slice-by-slice 2D U-net approach (second), the 3D U-net (third) and the  random 2.5D U-net (fourth).}
		\label{fig:fig2}
	\end{subfigure}\hspace{0.03\textwidth}
	\begin{subfigure}{0.482\textwidth}\vspace{2.33cm}
		\centering
		\includegraphics[width=0.315\textwidth]{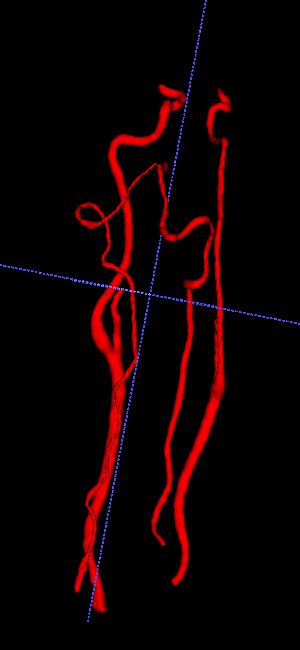}
		\hfil
		\includegraphics[width=0.315\textwidth]{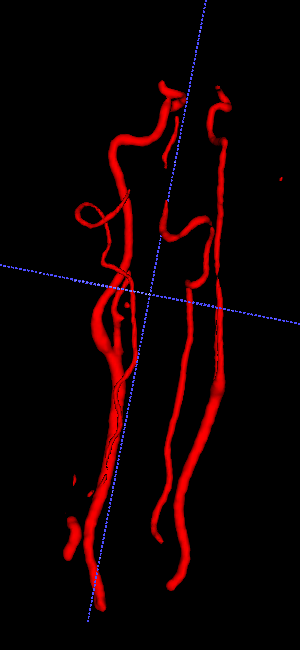}
		\hfil
		\includegraphics[width=0.315\textwidth]{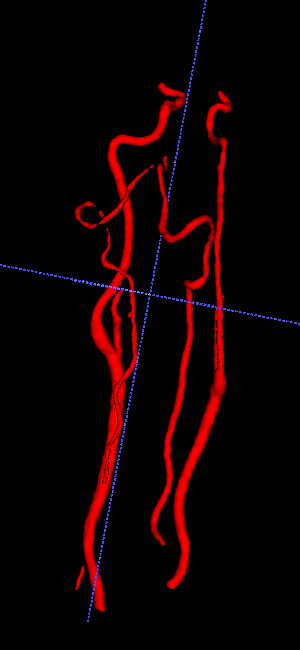}
		\newsubcap{Ground truth (left), 3D segmentation mask generated by the 3D U-net (middle) and by the random 2.5D U-net (right).}
		\label{fig:fig1}
	\end{subfigure}
\end{figure}

Although the 3D U-net  demonstrates   high precision   in  our  sparse application (Fig. \ref{fig:fig2},\ref{fig:fig1}), it produces too much artefacts, which constitutes an essential drawback. Additionally, we are very limited in the choice of convolution layers and  corresponding channel sizes due to the huge size of the input data. So with this model it is hardly possible to conduct end-to-end  segmentation for even larger biomedical scans with more channels on our GPU.

As we observe in Fig. \ref{fig:fig2},  the slice-by-slice approach even generates more artefacts. Also the high standard deviation  of the metrics between different test samples shows, that this approach does  not generate segmentations  with consistent reliability, which is a huge drawback.

Considering evaluation metrics, the random approach outperforms all prior discussed models and the problem of  artefacts vanishes (Fig. \ref{fig:fig2},\ref{fig:fig1}). Furthermore, it uses nearly the same memory space as the 3D convolution model during training, but is also  applicable to bigger input scans without memory concern due to the 2D convolution part. This is not  possible for the 3D U-net.
%
%
%

\section{Conclusion}
Although the construction of 3D segmentation masks with the help of 3D U-net \cite{OC16,EB17,vnet} delivers very satisfying results for vessel segmentation, generation of artefacts and the restrictions to model's complexity due to memory issues are hardly sustainable. Therefore, we proposed the random 2.5D U-net structure, that is able to conduct volumetric segmentation of very big biomedical 3D scans so we can train a network without any concern about memory space and input size. For the targeted application, the random 2.5D U-net even outperformed the standard slice-by-slice and 3D convolution approaches and showed more consistent accuracy for test application.

Note that the proposed network structure can be used for different projection techniques. For blood vessel segmentation, maximum intensity projection seemed to be the most fitting projection method due the high sparsity and orientation of the vessels. For volumetric end-to-end  segmentation of  different types of biomedical scans, other projection types can be used, e.g. the Radon-transform  for kidney segmentation using the corresponding inversion formula \cite{radon}.

There already exist approaches for saving memory and time in 3D convolution models using convolution with cross-hair filters \cite{vesselnet}. In future work, we will compare our methodology to this technique.  We will also  investigate the use of the random 2.5D U-net architecture for other projection types and  applications, e.g. for 3D multilabel segmentation of spines or volumetric image reconstruction.

%
%
\bibliographystyle{splncs04}
\bibliography{ref}
\end{document}